\documentclass{article}

\usepackage[preprint]{neurips_2026}

\usepackage[utf8]{inputenc}
\usepackage[T1]{fontenc}

\usepackage{float}
\usepackage{hyperref}
\usepackage{url}
\usepackage{booktabs}
\usepackage{amsfonts}
\usepackage{amsmath}
\usepackage{nicefrac}
\usepackage{microtype}
\usepackage{xcolor}
\usepackage{graphicx}
\floatstyle{ruled}
\newfloat{algorithm}{tbp}{loa}
\floatname{algorithm}{Algorithm}

\newcommand{\FailureOneMean}{6.50}

\newcommand{\LogitGapThreeMean}{12.17}

\newcommand{\LenOneFrac}{28.1\%}
\newcommand{\LenOneTokenFrac}{4.3\%}
\newcommand{\FailureTwoLenOneFrac}{4.1\%}

\newcommand{\FwNumDocs}{2\,million}
\newcommand{\FwFailureOneMean}{9.56}

\newcommand{\LambadaFTwoRate}{17.4}
\newcommand{\LambadaFTwoCode}{15.4}
\newcommand{\LambadaFTwoShuf}{13.5}
\newcommand{\LambadaFTwoRound}{64.5}
\newcommand{\HotpotFTwoRate}{13.5}
\newcommand{\HotpotFTwoCode}{17.0}
\newcommand{\HotpotFTwoShuf}{18.0}
\newcommand{\HotpotFTwoRound}{27.0}

\newcommand{\ReconstructionTable}{%
  \begin{table*}[ht]
    \caption{Single-code reconstruction on 500 held-out Wikipedia windows,
      Transformer encoder (left) versus SONAR (right); both use 1024-dimensional
      codes. Exact is full-window exact match; RL is ROUGE-L; suffix is the mean
      length of the longest exactly reconstructed suffix --- the contiguous run of
      most-recent tokens the backward decoder emits without an error. }
    \label{tab:single-recon}
    \centering
    \begin{minipage}[t]{0.49\textwidth}
      \centering
      \small
      \setlength{\tabcolsep}{3pt}
      \textbf{ReconSpan Transformer}\\[2pt]
      \begin{tabular}{@{}rrrrrr@{}}
        \toprule
        $\ell$ & Exact & BLEU & RL   & F1   & Suffix \\
        \midrule
        8      & .700  & .879 & .942 & .945 & 6.47   \\
        16     & .452  & .844 & .921 & .927 & 10.86  \\
        32     & .022  & .563 & .735 & .777 & 9.94   \\
        64     & .000  & .281 & .465 & .559 & 8.15   \\
        128    & .000  & .160 & .312 & .405 & 8.26   \\
        256    & .000  & .087 & .215 & .283 & 7.42   \\
        \bottomrule
      \end{tabular}
    \end{minipage}\hfill
    \begin{minipage}[t]{0.49\textwidth}
      \centering
      \small
      \setlength{\tabcolsep}{3pt}
      \textbf{SONAR}\\[2pt]
      \begin{tabular}{@{}rrrrrr@{}}
        \toprule
        $\ell$ & Exact & BLEU & RL   & F1   & Suffix \\
        \midrule
        8      & .146  & .703 & .804 & .806 & 5.00   \\
        16     & .066  & .722 & .829 & .832 & 7.94   \\
        32     & .008  & .668 & .812 & .820 & 7.24   \\
        64     & .000  & .454 & .667 & .699 & 1.16   \\
        128    & .000  & .167 & .378 & .480 & 0.05   \\
        256    & .000  & .060 & .209 & .311 & 0.03   \\
        \bottomrule
      \end{tabular}
    \end{minipage}
  \end{table*}%
}

\newcommand{\MTEBTable}{%
  \begin{table}[H]
    \caption{MTEB is a standard benchmark that measures how well fixed-size text
      embeddings capture meaning, through semantic-similarity and retrieval tasks; here
      we apply it to the raw codes. STSBenchmark and STS17 (English) score how well code
      cosine similarity tracks human sentence-pair ratings, and SICK-R does the same on
      sentence pairs probing compositional meaning. NFCorpus is a biomedical
      document-retrieval task. Higher is better on all four.}
    \label{tab:mteb}
    \centering
    \small
    \begin{tabular}{llrrrr}
      \toprule
      Task           & Metric   & Projected Mamba & Transformer & Mamba hidden & SONAR          \\
      \midrule
      STSBenchmark   & Spearman & .4478           & .4297       & .4495        & \textbf{.6739} \\
      STS17 (en--en) & Spearman & .1631           & .2038       & -.0771       & \textbf{.6360} \\
      SICK-R         & Spearman & .4761           & .4845       & .5079        & \textbf{.6294} \\
      NFCorpus       & NDCG@10  & .0203           & .0448       & .0232        & \textbf{.0556} \\
      \bottomrule
    \end{tabular}
  \end{table}%
}

\newcommand{\ChunkReconstructionTable}{%
  \begin{table}[ht]
    \caption{Native reconstruction under ReconSpan boundaries on held-out WikiText \citep{wikitext}.
      \emph{Mean chunk len.}\ is the mean chunk length in tokens.
      \emph{Exact chunk} is the fraction of chunks reconstructed perfectly;
      \emph{exact token} is the fraction of document tokens that lie inside such
      perfect chunks; \emph{suffix} is the fraction of tokens inside each chunk's
      correctly reconstructed newest-first run, crediting chunks that are only
      partly right. PPL is conditional Qwen2.5-1.5B perplexity \citep{qwen25}. The \emph{random}
      row places boundaries uniformly while matching \emph{Failure(1)}'s per-document
      chunk count; despite similar exact-chunk accuracy, it trails ReconSpan on all
      token-level measures.}
    \label{tab:chunk-recon}
    \centering
    \small
    \begin{tabular}{lrrrrr}
      \toprule
      Stop rule           & Mean chunk len. & Exact chunk & Exact token & Suffix & PPL   \\
      \midrule
      \emph{Failure(1)}   & 6.50            & .797        & .907        & .944   & 15.95 \\
      random (matched)    & 6.50            & .794        & .618        & .767   & 20.03 \\
      \emph{Failure(2)}   & 12.28           & .526        & .536        & .740   & 18.43 \\
      \emph{Logit-gap(1)} & 8.31            & .730        & .809        & .890   & 16.84 \\
      \emph{Logit-gap(3)} & 12.17           & .553        & .541        & .715   & 20.37 \\
      \bottomrule
    \end{tabular}
  \end{table}%
}

\newcommand{\ReaderPPLTable}{%
  \begin{table}[ht]
    \caption{Conditional perplexity of generated continuations on 150 held-out
      WikiText windows under Qwen2.5-1.5B \citep{qwen25}; lower is better. The Llama reader and
      human continuations are similar, while the smaller reader and native decode
      are moderately higher. Learned readers use top-$k$ sampling; repetitive greedy
      decoding is omitted.}
    \label{tab:reader-ppl}
    \centering
    \small
    \begin{tabular}{lr}
      \toprule
      Continuation source       & PPL   \\
      \midrule
      Human continuation        & 11.95 \\
      Native autoencoder decode & 15.95 \\
      Pythia-410M reader        & 13.94 \\
      Llama-3-8B reader         & 11.78 \\
      \bottomrule
    \end{tabular}
  \end{table}%
}

\newcommand{\DownstreamTable}{%
  \begin{table}[t]
    \caption{Downstream readout from ReconSpan latent tokens (\%). \emph{Control}
      substitutes another example's codes. \emph{Text}
      gives raw context to the untouched base model, and \emph{roundtrip} gives it
      autoencoder-reconstructed text. Metrics are exact next-word accuracy for LAMBADA,
      four-way closed-set accuracy for AG News, and answer exact match for HotpotQA.
      A superscript $*$ marks readout above the shuffled control at 95\% confidence, using
      a normal approximation to the difference of binomial proportions with standard
      errors computed from the evaluation-set size.
      Task FT is a supervised direct-readout reference, not a matched-training baseline.
      Mean len. is the average number of input tokens per latent token; unmarked rows use
      \emph{Failure(1)}.}
    \label{tab:downstream}
    \centering
    \small
    \setlength{\tabcolsep}{4pt}
    \begin{tabular}{llrrrrr}
      \toprule
      Task     & Reader / route                & Mean chunk len.    & Readout            & Control            & Text & Roundtrip           \\
      \midrule
      LAMBADA  & Pythia-410M                   & 10.2               & 5.1$^*$            & 1.8                & 51.0 & 50.2                \\
               & Llama-3-8B                    & 10.2               & 10.0$^*$           & 5.5                & 75.8 & 72.5                \\
               & Llama-3-8B + task FT          & 10.2               & 20.1$^*$           & 6.2                & 75.8 & 72.5                \\
               & Llama-3-8B, \emph{Failure(2)} & \LambadaFTwoRate{} & \LambadaFTwoCode{} & \LambadaFTwoShuf{} & 75.4 & \LambadaFTwoRound{} \\
      \midrule
      AG News  & Pythia-410M                   & 7.7                & 52.5$^*$           & 26.4               & 53.0 & 52.8                \\
               & Llama-3-8B                    & 7.7                & 74.3$^*$           & 24.5               & 72.6 & 73.0                \\
               & Llama-3-8B + task FT          & 7.7                & 82.7$^*$           & 24.2               & 72.6 & 73.0                \\
               & Llama-3-8B, \emph{Failure(2)} & 12.9               & 74.6$^*$           & 25.6               & 72.1 & 72.0                \\
      \midrule
      HotpotQA & Pythia-410M                   & 7.5                & 0.0                & 1.0                & 7.0  & 6.0                 \\
               & Llama-3-8B                    & 7.5                & 17.0               & 18.0               & 36.0 & 33.0                \\
               & Llama-3-8B + task FT          & 7.5                & 30.0               & 23.0               & 36.0 & 33.0                \\
               & Llama-3-8B, \emph{Failure(2)} & \HotpotFTwoRate{}  & \HotpotFTwoCode{}  & \HotpotFTwoShuf{}  & 36.0 & \HotpotFTwoRound{}  \\
      \bottomrule
    \end{tabular}
  \end{table}%
}

\title{ReconSpan: Reconstruction-Guided Adaptive Latent Tokenization}
\workshoptitle{Linguistic Principles for Foundation Models}

\author{%
  Lixing Li \\
  Cornell University \\
  Ithaca, NY 14853, USA \\
  \texttt{ll963@cornell.edu} \\
}

\hypersetup{
  pdftitle={ReconSpan: Reconstruction-Guided Adaptive Latent Tokenization},
  pdfauthor={Lixing Li}
}

\begin{document}

\maketitle

\begin{abstract}
  Adaptive latent tokenization maps a fine-grained input to a shorter sequence of
  continuous representations associated with input-dependent spans. We introduce
  \emph{ReconSpan}, which divides text into chunks that a backward decoder can reconstruct
  from a single contextual prefix code and retains one such code as the latent token for
  each chunk. The reconstruction criterion is applied when chunks are formed, allowing one
  trained autoencoder to produce average chunk lengths from 6.5 to
  12.2. At matched average length,
  reconstruction-guided boundaries preserve more text than random boundaries. Readers of
  the resulting latent sequence recover topic information reliably but struggle to extract
  exact details.
\end{abstract}

\section{Introduction}
\label{sec:intro}

Language models do not operate directly on raw text: tokenization determines the sequence
positions to which representation and computation are assigned. Conventional subword
tokenizers choose these units before the model runs, largely from corpus-level frequency
statistics \citep{bpe}. Their granularity is therefore fixed rather than conditioned on
the actual input.

Byte- and character-level inputs avoid a fixed subword vocabulary and retain fine-grained
information, but they also produce substantially longer sequences \citep{charformer,blt}.
Adaptive latent tokenization instead groups neighboring units into input-dependent spans
and represents each span with one continuous latent token. Its chunking rule is central:
boundaries determine when new latent positions are created and thus where representation
and computation are allocated. Here, a latent token is a contextual representation assigned
to a span boundary; unlike a conventional token embedding, it may encode preceding context
as well as the associated span. This makes unit formation a model-dependent allocation
decision rather than a fixed preprocessing choice.

We propose \emph{ReconSpan}, an adaptive latent-tokenization method that divides text into
chunks a decoder can reconstruct from a single prefix code and retains one such code as the
latent token for each chunk. A
causal encoder produces a code at every position; starting from the final code, a backward
decoder reconstructs until an error criterion is exceeded, accepts the resulting suffix as
a chunk, and repeats from the first unreconstructed position. Differences in reconstruction
reach give shorter chunks to difficult spans and longer chunks to easier ones. Because the
criterion is applied only during chunking, it can also be relaxed to increase average chunk
length without retraining, giving ReconSpan input-dependent allocation and post-training
control of granularity.

We evaluate both the induced tokenization and the information accessible from its latent
tokens. ReconSpan produces average chunk lengths from 6.5 to 12.2 tokens, and
at matched average length its reconstruction-guided boundaries preserve more text than
random boundaries. Native reconstruction tests the selected spans through the autoencoding
route, while a separately trained \emph{reader} consumes the contextual latent tokens
directly and predicts text or task outputs. These readers recover topic information
reliably but struggle to extract exact details, exposing a gap between information retained
by the autoencoder and information accessible to another model.

Our contributions are:
\begin{itemize}
  \item We introduce reconstruction fidelity as a chunk-allocation criterion for adaptive
        latent tokenization over an existing subword sequence.
  \item We show that one autoencoder supports multiple post-training granularities and that
        its variable-span boundaries outperform length-matched random boundaries in native
        reconstruction.
  \item We characterize the resulting contextual latent tokens through direct readout,
        separating information retained by the autoencoder from information accessible to
        readers of different scales and levels of task adaptation.
\end{itemize}

\section{Related Work}
\label{sec:related}

Learned tokenizers and hierarchical sequence models differ in the signal that determines
their source-text chunks. We organize the closest work by this allocation rule.

\paragraph{Chunking by fixed position.}
MEGABYTE divides byte sequences into fixed-size patches and applies separate local and
global models within and across patches \citep{megabyte}. Extensible Tokenization instead
contextualizes existing subword embeddings and retains representations at a regular stride
\citep{extensible}. Its stride can be selected at inference, but equal-size allocation does
not adapt boundaries to the input. It is especially close to ReconSpan in accepting
subword inputs, emitting continuous contextual representations, and allowing granularity
to change after training.

\paragraph{Chunking by predictive entropy.}
The Byte Latent Transformer (BLT) creates variable byte patches at spikes in next-byte
entropy, assigning shorter patches where the sequence is less predictable \citep{blt}.
Dynamic Token Pooling also studies an entropy-supervised boundary predictor alongside its
other variants \citep{dtp}. These methods use forward predictive uncertainty; ReconSpan
instead measures backward reconstruction fidelity.

\paragraph{Chunking by semantic similarity.}
SemToken embeds existing tokens contextually, merges adjacent semantically similar spans,
and varies granularity with local semantic density \citep{semtoken}. This directly targets
semantic allocation. ReconSpan does not optimize similarity or claim that its boundaries
are semantic; whether reconstruction difficulty aligns with linguistic structure remains
an empirical question.

\paragraph{Chunking by learned boundary scores.}
Dynamic Token Pooling predicts variable character-level segments using end-to-end,
tokenizer-supervised, entropy-supervised, or linguistic objectives \citep{dtp}. H-Net learns
content- and context-dependent routing jointly with a hierarchical byte-level language
model \citep{hnet}. Charformer is an earlier soft-block precursor: it scores candidate byte
blocks from the end-task loss, although its final downsampling is fixed \citep{charformer}.
FLEXITOKENS likewise learns variable byte boundaries while relaxing the fixed target-rate
objective used by related models \citep{flexitokens}. In these systems, allocation is
trained as part of language modeling.

\paragraph{Chunking by reconstruction fidelity.}
ReconSpan places a boundary according to how far a backward decoder can successfully
reconstruct from each contextual encoder state. The signal is measured autoencoder
reconstruction rather than a fixed position, learned router, forward entropy, or semantic
density. Changing the accepted reconstruction criterion adjusts average span length after
training. Our contribution is this allocation criterion and a characterization of the
resulting latent representations, not continuous tokens or adaptive segmentation by
themselves.

\section{Method}
\label{sec:method}

ReconSpan requires a generic autoencoding model and an inference-time chunking algorithm.

\subsection{Autoencoding model}
\label{sec:method-model}
Let $x_{1:n}$ be a token sequence. ReconSpan requires two learned components.
\begin{itemize}
  \item A \emph{prefix encoder} maps any token sequence to one code,
        \begin{equation}
          E:\mathcal{V}^{*}\rightarrow\mathbb{R}^{d},\qquad c_t=E(x_{1:t}).
        \end{equation}
        A causal model such as a Transformer or a Mamba can generate
        $c_1,\ldots,c_n$ in one pass.
  \item A \emph{backward decoder} maps $c_t$ to the encoded tokens in reverse order,
        $x_t,x_{t-1},\ldots$. Backward decoding is the point of the design: decoding
        newest-first, the position where the decoder first fails is a direct measurement of
        how far back that one code reconstructs. A forward decoder would instead have to be
        told where to start --- the very quantity we want to measure. ReconSpan's
        reach-based criterion therefore relies on backward decoding.
\end{itemize}
The autoencoding path is therefore
\begin{equation}
  (x_1,x_{2},\ldots,x_t)\xrightarrow{E}c_t\xrightarrow{D}(x_t,x_{t-1},\ldots,x_1).
\end{equation}

\subsection{Chunking algorithm}
\label{sec:method-chunking}
The decoder will not reconstruct arbitrarily long prefixes, so we only ask it to decode
within its own capacity; the point at which it fails sets each chunk boundary, giving
adaptive-length chunks and their latent tokens. Algorithm~\ref{alg:reconspan} states the
procedure. Because the text being tokenized is already known, the decoder is
\emph{teacher-forced} against it: at each reverse step it is fed the true previous tokens and
we record only whether its own greedy (argmax) prediction matches --- nothing is sampled or
generated.

Write the resulting chunk endpoints in chronological order as
$0=b_0<b_1<\cdots<b_m=n$, so chunk $i$ is $x_{b_{i-1}+1:b_i}$ and its contextual
latent token is $c_{b_i}=E(x_{1:b_i})$.

\begin{algorithm}[h]
  \caption{ReconSpan chunking.}
  \label{alg:reconspan}
  \textbf{Input:} tokens $x_{1:n}$, encoder $E$, backward decoder $D$, stopping rule.\quad
  \textbf{Output:} contextual latent-token sequence $\mathcal{C}$.
  \begin{enumerate}\setlength{\itemsep}{2pt}\setlength{\parskip}{0pt}
    \item Compute all prefix codes $(c_1,\ldots,c_n)\gets E(x_{1:n})$ in one causal pass;
          set $\mathcal{C}\gets\langle\,\rangle$ and $t\gets n$.
    \item \textbf{while} $t\geq 1$:
          \begin{enumerate}\setlength{\itemsep}{2pt}
            \item Teacher-force $D$ from $c_t$ against $x_t,x_{t-1},\ldots$ until the stopping rule
                  fires at $x_b$, so that $x_{b+1},\ldots,x_t$ are accepted; set $b\gets0$
                  if it reaches the beginning without firing. If it fires on the first step,
                  set $b\gets t-1$ (a one-token chunk, guaranteeing progress).
            \item Append $c_t$ to $\mathcal{C}$ as the code for chunk
                  $x_{b+1},\ldots,x_t$; set $t\gets b$ to resume from the first
                  unreconstructed endpoint.
          \end{enumerate}
    \item Reverse the collected latent tokens into chronological order and \textbf{return}
          $\mathcal{C}=(c_{b_1},\ldots,c_{b_m})$ --- only these chunk-boundary
          prefix codes are retained.
  \end{enumerate}
\end{algorithm}

We use two stopping families. \emph{Failure($m$)} stops at the $m$th incorrectly reconstructed token.
\emph{Failure(1)} ends a chunk at the first mistake and larger $m$ tolerates errors.
\emph{Logit-gap($\tau$)} is the continuous version. Let $y_j$ be the actual token at reverse
step $j$; the decoder stops at the first $k$ for which
\begin{equation}
  G_k=\sum_{j=1}^{k}\left(\max_{v\in\mathcal V}z_v^{(j)}-z_{y_j}^{(j)}\right)>\tau .
  \label{eq:logitgap}
\end{equation}
Each summand is zero when the model predicts the correct token and otherwise
measures by how much it was missed, so $G_k$ accumulates near-misses instead of counting
outright errors. Raising $m$ or $\tau$ lengthens chunks on a fixed trained model and thus
increases the average number of input tokens represented by each latent token.

The number of sequential model invocations, which is architecture-invariant, is $O(1+n)$.
Producing all prefix codes $c_1,\ldots,c_n$ takes a single encoder forward pass, whatever the
encoder architecture. The backward reconstruction is autoregressive, so in the worst case ---
a failure at every step, making every chunk one token --- it needs $O(n)$ sequential decoder
calls.

In practice, the decoder reads a fixed block of $W$ positions per
endpoint for efficient batching, so decoding is $O(Wn)$ work in the worst case with our specific Mamba decoder.
Appendix~\ref{app:dp} describes a variant that keeps the same $O(Wn)$ total work but cuts the
sequential decode calls to $O(W)$, independent of $n$, by materializing all endpoint decodes
at once, at the cost of more physical computation.

\subsection{Training}
\label{sec:method-training}
The only training objective is autoencoder reconstruction; the span allocation emerges
from the decoder's capacity rather than from an explicit length target. A training
step operates on one window $x_{1:L}$. The encoder produces its final code
$c_L=E(x_{1:L})$, a learned projection maps it to the decoder's initial recurrent state,
and the decoder is teacher-forced to reproduce the window in reverse,
$y=(x_L,\ldots,x_1,\mathrm{EOS})$. Each step reconstructs the first $k\leq L+1$ reversed
targets, and the loss is the token cross-entropy over them,
\begin{equation}
  \mathcal L_{\mathrm{AE}}=-\sum_{j=1}^{k}
  \log p_D(y_j\mid y_{<j},c_L).
\end{equation}
The projection is differentiable and no stop-gradient is placed on the code, so this loss
reaches the encoder as well as the decoder. Encoder and decoder can therefore be trained
jointly.

Capping the decode length at $k\leq L+1$ makes most steps reconstruct only a \emph{suffix}
of the window rather than the whole text. This is a deliberate match to how ReconSpan
uses the decoder: the design is intended to prioritize recent tokens and thereby extend
the successful suffix that determines a chunk boundary.

\subsection{Implementation overview}
\label{sec:method-implementation}
Our encoder is a Pythia-410M Transformer \citep{pythia} whose final 1024-dimensional hidden
state is the code; the decoder is a Mamba2-130M backward model \citep{mamba2}. The code is
up-projected to the decoder's per-layer initial SSM states, and a BOS token begins the
rollout. Pythia shares the GPT-NeoX tokenizer with the Mamba2 decoder, so the two never
disagree on vocabulary. We also evaluate two auxiliary Mamba2 encoders: a 4096-dimensional
projected SSM state and a 1024-dimensional hidden state. The former requires four times the
code width to approach the Transformer's short-span reconstruction, while the latter is
weaker at equal width. Appendix~\ref{app:impl} defines these
variants and reports their reconstruction comparison.

Each step samples a short-biased encode length $L$ and an independent decode length
$k\leq L+1$; Section~\ref{sec:method-training} explains why a partial suffix, not the full
window, is the right target. Training runs in two stages on FineWeb \citep{fineweb}: a
7B-token backward-decoder pretrain with the encoder frozen, then 3B tokens of joint
encoder--decoder training. Appendix~\ref{app:impl} gives the data pipeline, sampling ranges,
schedule, and optimizer.

\section{Tokenizer properties}
\label{sec:tokenizer-properties}

We characterize the tokenizer through four measurements: single-code autoencoder
reconstruction, semantic structure in the raw code geometry, the span lengths induced by
different stopping rules, and how much text survives a native autoencoder round trip under
the selected boundaries.

\subsection{Autoencoder reconstruction quality}
\label{sec:single-code}

\ReconstructionTable

We measure how much a single code can give back, comparing our Transformer encoder against
SONAR \citep{sonar}. For each length $\ell$ we sample one $\ell$-token window from each of
500 held-out Wikipedia documents, encode the whole window into one code, decode it
autoregressively in reverse, and score the result against the original using BLEU
\citep{bleu} and ROUGE-L \citep{rouge}, alongside exact-match measures. SONAR is a
multilingual text autoencoder that maps a sentence to one 1024-dimensional embedding and
decodes it back, so its code width matches ours; as a strong open-source autoencoder it
gives an external reference point for our reconstruction quality.

Table~\ref{tab:single-recon} reports the comparison. The key column is \emph{suffix}, the
quantity most closely related to the reconstruction reach used to set chunk length. The
Transformer recovers an average exact suffix of roughly $8$--$11$ tokens from $\ell=8$ to
$\ell=256$, whereas SONAR's suffix falls sharply beyond $\ell=32$ even when its aggregate
overlap remains competitive. This stable short-range reconstruction is the capacity needed
by the chunking experiments that follow.

\subsection{Semantic geometry}
\label{sec:semantic-geometry}
Reconstruction asks a code to retain the tokens of its window, but does not directly place
semantically similar inputs nearby. We probe this untrained property with four tasks from
MTEB \citep{mteb}: STSBenchmark and STS17 \citep{cer2017sts}, SICK-R
\citep{marelli2014sick}, and NFCorpus \citep{boteva2016nfcorpus}. Table~\ref{tab:mteb}
shows that SONAR, which is trained with a similarity objective, leads on every task, often
by a wide margin. Thus the raw code geometry is not strongly organized by semantic
similarity. This protocol evaluates one full-input encoder code rather than a sequence of
boundary tokens, so it characterizes the code generator rather than the complete
tokenizer.

\MTEBTable

\subsection{Adaptive span allocation}
\label{sec:rate}

\begin{figure}[ht]
  \centering
  \includegraphics[width=0.86\linewidth]{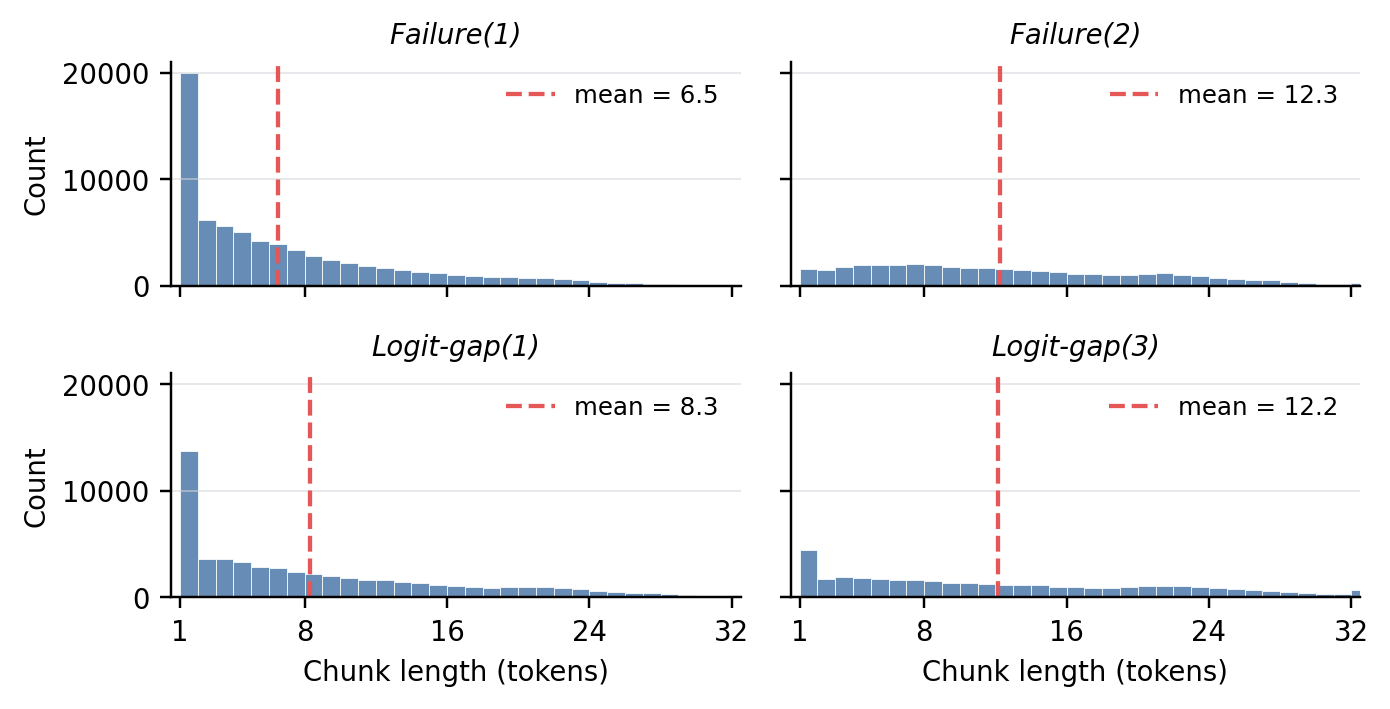}
  \caption{Chunk-length distributions (raw counts) for four stopping rules on the same 500
    Wikipedia documents, each capped at 1000 tokens; vertical lines mark the mean. Looser
    rules (left to right, top to bottom) shift mass toward longer chunks. Teacher-forced
    decoding evaluates $W=32$ positions per endpoint, so the $W{=}32$ cap truncates the few
    windows the decoder could carry further, producing the small pile-up in the length-32
    bin (<$1\%$ of chunks).}
  \label{fig:chunk-distribution}
\end{figure}

We study how the stopping criterion controls span allocation and whether the resulting
chunk lengths vary across inputs. Because every chunk contributes one latent position,
mean chunk length directly measures the tokenizer's average granularity. A looser rule ---
one that permits the decoder to continue further before stopping --- increases this mean.
Figure~\ref{fig:chunk-distribution} shows this across four rules applied post hoc to one
trained model on 500 Wikipedia documents, spanning mean lengths of \FailureOneMean{} to
\LogitGapThreeMean{} input tokens. The behavior carries to larger scale:
over \FwNumDocs{} FineWeb documents, \emph{Failure(1)} gives a mean chunk length of
\FwFailureOneMean{} tokens. Realized averages also differ across Wikipedia, FineWeb, and
the downstream datasets in Section~\ref{sec:downstream}, reflecting input-dependent
variation in reconstruction reach.

The one caveat is the spike at length one, which has a mechanical cause. The backward decoder
predicts the newest token first, with no reconstructed context to condition on, so that
token is the hardest to get right. Two outcomes then both produce a length-one chunk: the
decoder misses this first token, and the forced-progress rule of
Section~\ref{sec:method-chunking} retains it anyway; or it reconstructs the first token
but misses the second. \emph{Failure(1)} thus piles both the stop-at-one and stop-at-two cases
onto length one. These single-token chunks are common but carry only \LenOneTokenFrac{} of
the text, so they cost latent positions without greatly changing the mean; a looser rule or an
explicit minimum chunk length removes most of them.

\subsection{Native reconstruction under selected boundaries}
\label{sec:native-reconstruction}
We test whether capacity-based boundaries identify spans that survive a native
autoencoder round trip. After ReconSpan selects the boundaries, each resulting chunk is
encoded in isolation and decoded autoregressively; Appendix~\ref{app:roundtrip} defines
this protocol formally and distinguishes its chunk codes from the prefix codes consumed
directly by the reader.

Two conclusions follow. First, at an identical latent-token count and mean length,
ReconSpan boundaries recover more of each document than the length-matched random control,
so where the boundaries fall, not merely how many there are,
decides how much text survives. Second, the stopping rule acts as a granularity dial. The
results suggest that mean chunk length largely predicts quality across the two rule
families: failure and logit-gap rules at similar mean lengths reach similar reconstruction
quality. Thus the same trained model exposes a controllable quality--granularity tradeoff.

\ChunkReconstructionTable

\section{Reading the latent tokens}
\label{sec:readout}

Can downstream language models be trained to operate directly on ReconSpan's
variable-length latent-token sequence, and what information can they recover? We first
define and train readers that predict text directly from the latent tokens, then evaluate
their access to semantic, lexical, and retrieval information.

\subsection{Reader model and training}
\label{sec:reader-architectures}
We train a separate language model, called a \emph{reader}, to predict text directly from
the latent-token sequence rather than from the source text. For the boundaries
$0=b_0<\cdots<b_m=n$ from Section~\ref{sec:method-chunking}, the latent-token prefix at
boundary $b_i$ is
\begin{equation}
  \mathcal C_{\leq i}:=(c_{b_1},\ldots,c_{b_i}),\qquad c_{b_j}=E(x_{1:b_j}).
\end{equation}
The reader $R$ maps this variable-length sequence of continuous codes to a distribution
over output token sequences,
\begin{equation}
  R:(\mathbb R^d)^*\rightarrow\Delta(\mathcal V^*),\qquad
  p_R(y\mid\mathcal C_{\leq i})=
  \prod_{j=1}^{|y|}p_R(y_j\mid\mathcal C_{\leq i},y_{<j}).
\end{equation}
Thus the reader predicts a continuation or task output directly from contextual prefix
codes, without first decoding them back to text.

Each code is standardized coordinate-wise using corpus-level mean and standard deviation,
then mapped to the reader's embedding dimension by a learned linear adapter and RMSNorm
\citep{rmsnorm}. The adapted codes are followed by a separator and the text output. During
generic training, the reader receives a randomly truncated latent-token prefix formed with
\emph{Failure(1)}; next-token loss is masked on the codes and separator and applied only to
the continuation. We fully fine-tune Pythia-410M, whereas Llama-3-8B \citep{llama3} updates
LoRA weights \citep{lora}, the adapter, and RMSNorm. Training uses generic FineWeb
continuations.

Table~\ref{tab:reader-ppl} shows that the latent tokens support fluent generation. Under an
external perplexity scorer, continuations from the larger reader are comparable to human
text. Fluency does not establish faithfulness, however: qualitative samples preserve
topic, register, and local syntax while exact content remains difficult to access.
Appendix~\ref{app:lcm} gives further implementation details and reports unsuccessful
code-to-code reader variants.

\ReaderPPLTable

\subsection{Information accessible to downstream readers}
\label{sec:downstream}
We test what the reader can recover along a spectrum of information specificity. AG News
\citep{agnews} tests coarse semantic information through four-way topic classification.
LAMBADA \citep{lambada} requires exact lexical information to predict a passage's final
word. HotpotQA \citep{hotpotqa} requires multi-hop retrieval from ten documents followed
by open-answer generation. Across these tasks, we vary reader scale, task-specific
adaptation, and chunking granularity.

Table~\ref{tab:downstream} measures direct access against two references. The shuffled
control replaces each example's codes with codes from another example, so improvement over
it shows that predictions use example-specific information. The round trip instead decodes
isolated chunk codes back to text before prediction; it measures information retained by
the native autoencoder route and provides a route-specific reference. Together, these
comparisons separate information retention from reader access.

Across the three tasks, readers access coarse semantics more readily than exact details.
The Llama reader reaches raw-text-level topic classification on AG News, whereas LAMBADA
remains far below the round-trip reference. Generic readout is also weak on HotpotQA and
does not improve over the shuffled control. Thus the codes support semantic recognition,
but exact lexical access and evidence retrieval remain difficult for a generic reader.

Task adaptation provides a second, supervised reference for direct readout under the
training used here. It improves all three tasks and moves HotpotQA toward the round-trip
reference, although its advantage over the shuffled control does not reach the marked
confidence threshold. These gains show that targeted training can extract task-relevant
information left unused by generic FineWeb continuation training.

Although the readers are trained only with \emph{Failure(1)} codes, they do not collapse
when evaluated with the coarser \emph{Failure(2)} policy. Topic classification remains
stable, LAMBADA remains numerically above its shuffled control, and HotpotQA is comparable
to its generic \emph{Failure(1)} result. The learned interface therefore tolerates a
substantial change in chunk granularity without retraining.

Overall, the native route retains information that current readers do not fully access.
Topic information is readily recoverable by a downstream language model, task supervision
narrows the access gap, and exact-detail readout remains the central limitation.

\DownstreamTable

\section{Limitations and future directions}
\label{sec:limitations}

The experiments expose several limitations of the current tokenizer, autoencoder, and
reader.

\subsection{Limitations}
\paragraph{Reader access.}
Direct-code readers remain weak on exact-content and retrieval tasks, even when scaling
the reader or adapting it to the target task improves performance. By contrast, the
native round trip answers the same tasks much better, showing that isolated codes for the
selected chunks retain information the current reader cannot yet access. It is a reference
for the native reconstruction route, while task-adapted direct readout provides a separate,
supervised reference for reader access. One
hypothesis is that code geometry makes this access difficult: the autoencoder is trained
for reconstruction rather than to place semantically similar inputs nearby, and its weak
MTEB results (Section~\ref{sec:semantic-geometry}), especially on retrieval, are consistent with this
explanation but do not establish it as a cause.

\paragraph{Short chunks.}
Too much of the latent sequence is spent on very short chunks. Under \emph{Failure(1)}, \LenOneFrac{}
of chunks cover a single token. These chunks carry only
\LenOneTokenFrac{} of the text while consuming \LenOneFrac{} of the codes, so a
disproportionate share of the latent sequence represents very little source text. Relaxing the rule to
\emph{Failure(2)} reduces them to \FailureTwoLenOneFrac{}, which is part of why that setting
nearly doubles the mean chunk length at little measured cost on topic-level tasks; a stopping rule with
an explicit minimum chunk length is an obvious remaining improvement.

\paragraph{Backward-decoder pretraining.}
ReconSpan requires a model that decodes text backward from a code. Such pretrained
decoders are not readily available, so our decoder must be trained from scratch before the
autoencoder can be jointly optimized. This raises the entry cost relative to methods built
entirely from existing pretrained language models.

\paragraph{Boundary-selection cost.}
The reconstruction scan is computationally expensive at corpus scale. Although it is
$O(n)$ for fixed $W$, constructing the reader corpus took about 12 hours for 2M documents,
roughly two to three times the 4--5 hours of reader training it fed. Fewer latent positions
also do not by themselves establish lower end-to-end compute: the Transformer encoder has
quadratic work in source length, and a one-use setting must pay the boundary-selection cost
before any downstream benefit.

\paragraph{Scope of the tokenizer.}
ReconSpan operates over an existing subword sequence and is therefore a higher-level latent
tokenizer, not a replacement for byte-to-text tokenization. Its native reconstruction and
direct-reader routes also use different code constructions (Appendix~\ref{app:roundtrip}).
Finally, we do not compare downstream language-model quality or efficiency directly with
end-to-end byte tokenizers such as BLT or H-Net.

\subsection{Future directions}
Readers trained at larger scale, on better-matched data, or with objectives aimed at exact
retrieval may recover more of the information demonstrated by the native round trip.

Training the autoencoder with an additional semantic-clustering objective could make
related inputs easier for a reader to recognize. This would also directly test the
hypothesis, motivated by Section~\ref{sec:semantic-geometry}, that unstructured code geometry contributes to
weak readout.

A different native decoder could replace the backward language model. For example, a
Transformer decoder could receive the code through a soft prompt \citep{prompttuning} or
cross-attention \citep{transformer}.

The linguistic structure of the boundaries placed by ReconSpan remains unexplored.
Testing whether they align with syntax, discourse, or information density could explain
what the reconstruction criterion treats as difficult and guide better allocation rules.
Appendix~\ref{app:boundary} visualizes how changes in reconstruction reach produce the
boundaries for one example.

A further direction is to evaluate ReconSpan as a context-compression interface, measuring
prefill latency, KV-cache memory, boundary-selection cost, and amortization across repeated
reads rather than inferring efficiency from sequence reduction alone.

\section{Conclusion}
\label{sec:conclusion}
ReconSpan forms adaptive latent tokens by retaining contextual encoder states at boundaries
set by backward reconstruction reach. One autoencoder exposes mean chunk lengths from
\FailureOneMean{} to \LogitGapThreeMean{} after training, and its boundaries reconstruct
better than length-matched random ones. Downstream language models operate directly on
these sequences, recovering topics more readily than exact lexical details; task adaptation
narrows the native-reconstruction gap. These results establish reconstruction fidelity as
a viable allocation criterion and motivate stronger readers and linguistic boundary
analysis.

{
\small
\bibliographystyle{plainnat}
\bibliography{references}
}

\appendix

\section{Implementation and training details}
\label{app:impl}

\paragraph{Models and tokenizer.} All variants share a Mamba2-130M backward decoder
(24 layers) and the GPT-NeoX tokenizer. We evaluate three encoders: a \emph{projected SSM
  state} (Mamba2-130M, whose SSM state is compressed to a 4096-dimensional code), and two
\emph{hidden-state} encoders whose final 1024-dimensional hidden state is the code directly
--- Pythia-410M (Transformer) and Mamba2-370M. Pythia shares the GPT-NeoX tokenizer with the
Mamba2 checkpoints, so encoder comparisons introduce no vocabulary change.

\paragraph{Projections.} The two families project differently. For the hidden-state encoders
the code \emph{is} the encoder's final 1024-dimensional hidden state, so there is no
down-projection; a single learned linear maps the code to an
$n_{\text{layer}}\times k_c\times k_n$ tensor ($k_c{=}128$ channels, $k_n{=}8$ per layer),
which is expanded to each decoder layer's full SSM initial state
$(n_{\text{heads}},\text{headdim},N)=(24,64,128)$ by parameter-free \texttt{repeat\_interleave}
along the channel ($\times 12$) and state ($\times 16$) axes; convolution states start cold.
For the projected-SSM-state encoder the code is compressed \emph{from} the encoder's per-layer
states by a non-trivial map: each layer's state ($D{=}1536$ channels $\times\,N{=}128$ state)
is (i) channel-compressed by a learned linear $D\!\to\!k_c$, (ii) mean-pooled along the state
dimension $N\!\to\!k_n$ ($16\times$), and (iii) averaged over layers within four groups
$[(0,12),(12,18),(18,21),(21,24)]$; the four group codes are concatenated into the
$4\,k_c k_n=4096$-dimensional code. The lossy, parameter-free pooling is placed deliberately
on the state and layer axes, while the information-bearing channel dimension keeps a
learned map. The up-projection mirrors these
steps, and its weights are initialized as the pseudo-inverse of the down-projection so the
round trip is near-lossless before training.

\begin{figure}[t]
  \centering
  \includegraphics[width=\linewidth]{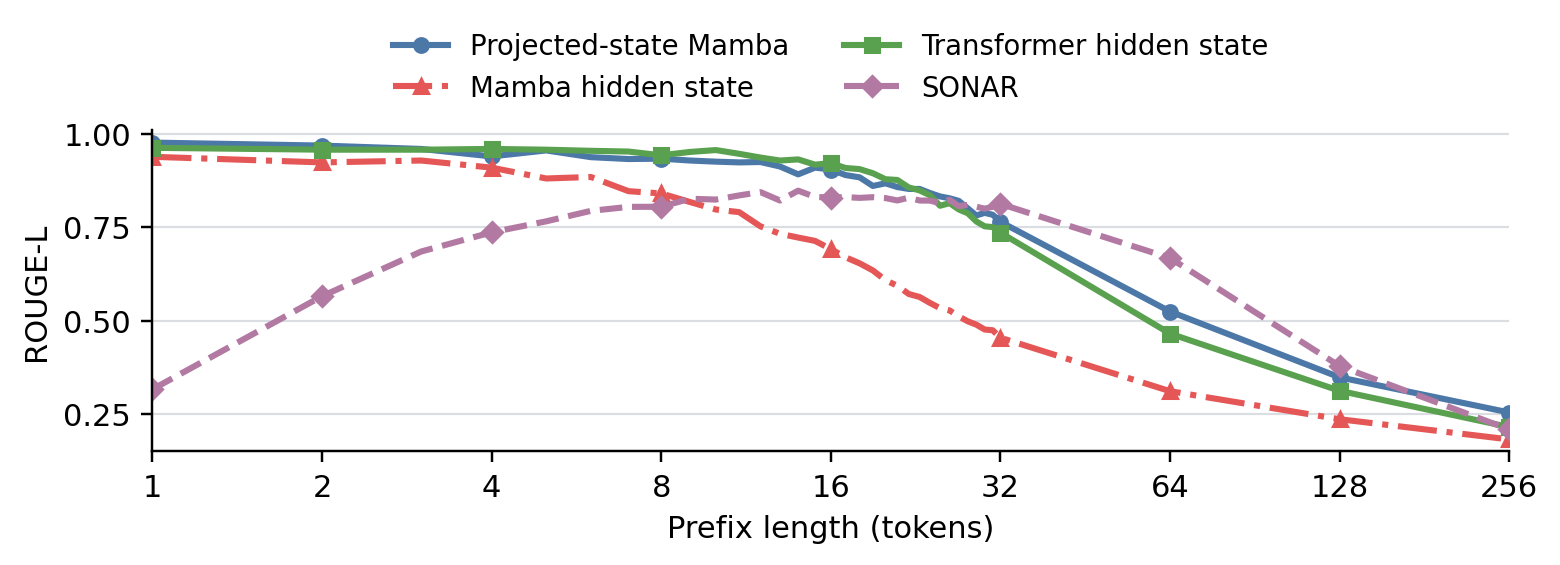}
  \caption{Single-code reconstruction ROUGE-L versus encode length, on 500 held-out
    Wikipedia windows (protocol in Section~\ref{sec:single-code}). Among the three encoders
    the Transformer hidden-state variant is strongest at the short lengths that set chunk
    boundaries; the projected SSM state matches it only at $4\times$ the code width, and the
    Mamba hidden state is weakest. SONAR is shown for reference; its drop below 16 tokens
    reflects training on full sentences rather than short fragments.}
  \label{fig:reconstruction}
\end{figure}

\paragraph{Why the widths differ.} A hidden state is 1024-dimensional, matched to SONAR's
width for the single-code comparison in Section~\ref{sec:single-code}. The full Mamba2-130M
SSM state is far larger --- 24 layers of $1536\times128$ values, about 4.7M in total --- and
already sparse, so compressing it to 1024 would discard too much; we keep 4096. Even at four
times the width, the SSM-state code reconstructs less well than the 1024-dimensional
Transformer code (Figure~\ref{fig:reconstruction}).

\paragraph{Encoder choice.} We evaluated all three encoders on the single-code protocol of
Section~\ref{sec:single-code} before committing to the Transformer.
Figure~\ref{fig:reconstruction} plots ROUGE-L against encode length for the three encoders
and SONAR. The Transformer hidden-state encoder is strongest at the short lengths
($\ell\le 32$) that set almost every chunk boundary; the projected SSM state matches it only
by spending the $4\times$ wider 4096-dimensional code, and the Mamba hidden state at equal
width falls off fastest. This is why the main paper uses the Transformer encoder throughout
and reports the other two only here.

\paragraph{Data and batching.} Training text is FineWeb, tokenized into a flat queue and
concatenated end-to-end with no separator, so a sampled window may cross a document boundary
(GPT-style). Each step samples one encode length $L$ from a mixture --- 25\% uniform on
$[1,20]$ and the remainder log-uniform on $[20,4096]$ --- and an independent decode length
$k$ log-uniform on $[1,4096]$ clipped to $k\leq L+1$. All samples in a step share the same
$L$, so each batch is a dense $(B,L)$ block that needs no padding; $B$ is set per step from a
throughput target and a length-aware memory cap.

\paragraph{Two-stage training.} Stage 1 freezes the encoder and trains the backward decoder
and projection for approximately 7B tokens. Stage 2 initializes each encoder variant from
that shared decoder, attaches a fresh projection, and jointly trains encoder, decoder, and
projection for a further 3B tokens.

\paragraph{Optimizer.} AdamW \citep{adamw} with encoder/decoder/projector learning rates
$5\!\times\!10^{-5}$, $10^{-4}$, and $10^{-3}$, weight decay 0.1, $\beta=(0.9,0.95)$, 1000
warmup steps, cosine decay to 10\% of peak, and gradient clipping at 1.0. The random seed is
42; checkpoints record model step and total tokens processed.

\paragraph{Compute.} All runs use a single NVIDIA A100 80GB PCIe with gradient checkpointing
\citep{gradientcheckpointing} enabled. Stage 1 (the shared 7B-token decoder pretrain) takes
about 28 hours; stage 2 (the
3B-token joint training) takes about 15 hours per encoder variant.

\section{Reproducibility and compute accounting}
\label{app:repro}
The implementation uses Linux, Python 3.10, CUDA 12.8, PyTorch 2.9.1, and Transformers
4.51.3. Random seeds are 42 for autoencoder training and data sampling, 0 for reader
validation splitting and task adaptation, and 0 for the rate-matched random-boundary
control. Source datasets and third-party model weights are obtained from the repositories
identified in Appendix~\ref{app:assets}. Code and checkpoints are not publicly released.

The downstream evaluations use 512 LAMBADA passages, 995 AG News articles, and the first
100 examples of the HotpotQA validation split. The confidence markers in
Table~\ref{tab:downstream} use these sample counts in the binomial standard errors stated
in its caption; they do not measure variation across training runs.

Table~\ref{tab:compute} accounts for the result-producing runs. Reader-corpus construction
is listed separately because it is a material boundary-selection cost rather than reader
optimization. The short task-adaptation runs took 12 minutes for LAMBADA (including 3.5
minutes of chunking), 5.3 minutes for AG News optimization, and 2.9 minutes for HotpotQA
optimization plus about 8 minutes of cached encoding. Other evaluation and plotting jobs
range from minutes to roughly two hours each. Including a conservative six GPU-hour
allowance for those jobs, the paper pipeline used approximately 94--102 A100 GPU-hours.
Retained job logs indicate that preliminary, ablation, debugging, and failed runs kept the
full research project below approximately 200 A100-equivalent GPU-hours. CPU-only dataset
loading and reporting were negligible relative to GPU work; the exact host CPU model was
not recorded. Peak local storage was about 140GB, dominated by cached reader data and
evaluation artifacts.

\begin{table}[ht]
  \caption{Compute for the principal paper pipeline on one NVIDIA A100 80GB PCIe. Ranges
    reflect measured variation between reader variants; task adaptation is itemized in
    the text.}
  \label{tab:compute}
  \centering
  \small
  \begin{tabular}{lrr}
    \toprule
    Component                                        & Runs & A100 hours \\
    \midrule
    Shared backward-decoder pretraining              & 1    & 28         \\
    Joint autoencoder training                       & 3    & 45         \\
    Two-million-document reader corpus               & 1    & 12         \\
    Generic Pythia and Llama readers                 & 2    & 8--10      \\
    Task adaptation and task-data encoding           & 3    & $<1$       \\
    Evaluation and plotting (conservative allowance) & ---  & $\leq 6$   \\
    \midrule
    Total for reported pipeline                      &      & 94--102    \\
    \bottomrule
  \end{tabular}
\end{table}

\section{Native autoencoder round trip}
\label{app:roundtrip}
The direct reader operates on cumulative prefix codes $\mathcal C_{\leq i}$ as defined in
Section~\ref{sec:reader-architectures}. The reconstruction experiment of
Section~\ref{sec:native-reconstruction} instead uses the same ReconSpan boundaries
to test each selected chunk through the native autoencoder. Given
$0=b_0<b_1<\cdots<b_m=n$, encode every chunk in isolation,
\begin{equation}
  k_i=E(x_{b_{i-1}+1:b_i}),\qquad
  \mathcal K=(k_1,\ldots,k_m).
\end{equation}
Let $D^*(k_i)$ denote the tokens produced autoregressively by the backward decoder before
EOS. Because the decoder generates newest-first, the reconstructed chunk and document are
\begin{equation}
  \hat x^{(i)}=\operatorname{reverse}(D^*(k_i)),\qquad
  \hat x=\hat x^{(1)}\Vert\cdots\Vert\hat x^{(m)},
\end{equation}
where $\Vert$ denotes concatenation. Boundary selection is teacher-forced, but this
round-trip decode conditions on its own previous outputs. Re-encoding isolated chunks is
used only for this native-decoding diagnostic and for decoded-text controls; the direct
reader path is simply $\mathcal C_{\leq i}\mapsto y$ and does not perform this extra work.

\section{Parallel boundary selection}
\label{app:dp}
Let $r_t\leq W$ be the backward reach obtained from the code ending at position $t$, where
$W$ is the decode block size of Section~\ref{sec:method-chunking}.
Greedy ReconSpan retains the current endpoint, moves to $t-r_t$, and repeats. Its work
is linear for fixed $W$, but the endpoint chain is serial and can contain $n$ one-token
chunks in the worst case.

The dynamic-programming variant evaluates the $W$ reverse positions for every endpoint
in parallel, producing all reaches $r_1,\ldots,r_n$. It then solves a minimum-cover
recurrence over candidate intervals $[t-r_t+1,t]$. Grouping transitions by reverse
offset requires $W$ parallel rounds after the prefix codes are available. The total work
remains $O(Wn)$ and the sequential decode calls become $O(1+W)$, but all $n$ endpoint decodes
are materialized rather than only the greedy boundary chain. This is faster in latency but
uses more physical compute and memory.

\section{Reader diagnostics and code-to-code reader details}
\label{app:lcm}
The code-to-code reader predicts the next \emph{chunk code} from the latent-token prefix,
following the Large Concept Model formulation \citep{lcm}. Using the boundaries
$b_1<\cdots<b_m$ found by ReconSpan, define the prefix code
$p_i=E(x_{1:b_i})$ and the independently re-encoded chunk code
$k_i=E(x_{b_{i-1}+1:b_i})$. The model is trained on sequences of these codes to predict
$k_{i+1}$ given prefix information through $p_i$.
Generation is a loop through the native decoder: predict a chunk code, decode it to text,
re-encode the generated text to update the prefix representation, and predict the next
code. The reader therefore depends on the code-to-text bridge as well as on its own
predictions.

Two variants were trained. A continuous diffusion head, which should model a multimodal
next-code distribution, collapsed toward the conditional mean and produced codes that did
not decode to usable text. Quant-LCM replaces the continuous target with residual vector
quantization --- the code is mapped to per-level codebook indices and predicted by
per-level softmax classification --- which trains stably and is the version reported here.
It uses an eight-layer, 1024-wide causal context tower over the chunk sequence. Its
conditional perplexity is 290.43 on the protocol of Table~\ref{tab:reader-ppl}, roughly
25 times the code-to-token reader's, and its generations are locally broken and
repetitive. We therefore report it as a negative result and use the code-to-token reader
in the main body. The failure is a prediction failure rather than a representation
failure: the same codes support fluent generation when the model is asked for tokens
instead of the next code.

For reference, the Pythia code-to-token reader fine-tunes all 406M base-model parameters
and a 1024-to-1024 adapter. Code positions are standardized and RMS-normalized; loss is
applied only to continuation tokens.

\section{Boundary observations}
\label{app:boundary}
\begin{figure}[H]
  \centering
  \includegraphics[width=\linewidth]{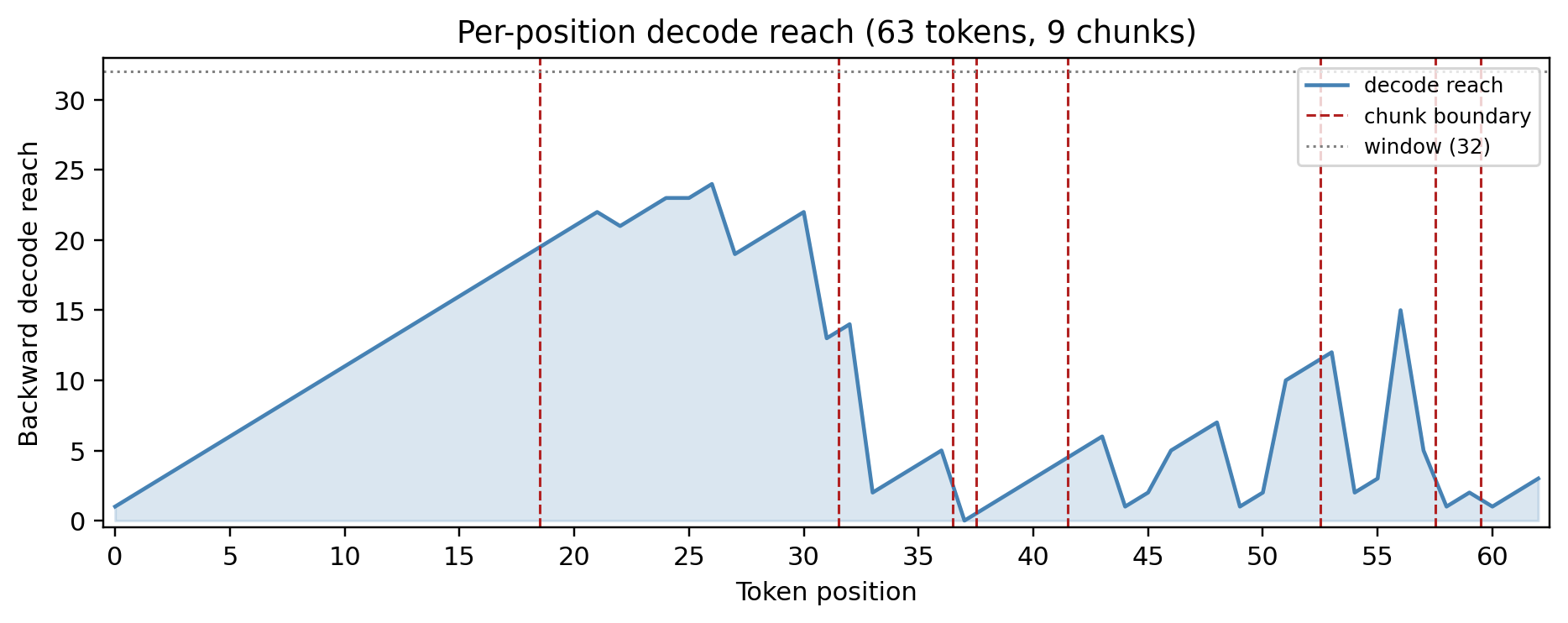}
  \caption{Per-position backward decode reach for one 63-token passage under
    \emph{Failure(1)}. Red dashed lines mark the eight interior boundaries of the nine chunks selected by ReconSpan,
    and the gray dotted line marks the 32-token decoding window. Reach changes sharply
    across positions, producing unequal chunk lengths and concentrating boundaries near
    positions where reconstruction fails early. This example motivates studying whether
    the boundaries align with syntax, discourse, or information density; it does not
    establish such alignment.}
  \label{fig:boundary-example}
\end{figure}

Figure~\ref{fig:boundary-example} shows how per-position reconstruction reach induces
unequal chunks in one document. We decode from every prefix code and record how many
consecutive tokens are reconstructed before the first error.
The gradual increases partly follow mechanically from
advancing the endpoint; the abrupt drops show positions from which the decoder fails much
earlier, forcing the greedy cover to allocate a new code. Whether such drops align
systematically with linguistic structure remains an open question.

The corresponding text is shown below, with vertical bars at the exact token boundaries
used in Figure~\ref{fig:boundary-example}.
\begin{quote}
  \small
  Water boils at 100 degrees Celsius, and Genghis Khan founded the Mongol
  \textcolor{red}{$\mid$} Empire in 1206. The clarinet has a single-reed
  \textcolor{red}{$\mid$} mouthpiece, while photos
  \textcolor{red}{$\mid$}ynthesis
  \textcolor{red}{$\mid$} converts sunlight into glucose
  \textcolor{red}{$\mid$}. Napoleon was exiled to Elba in 1814
  \textcolor{red}{$\mid$}, but the liver performs
  \textcolor{red}{$\mid$} over 500
  \textcolor{red}{$\mid$} metabolic functions.
\end{quote}
The within-word and punctuation-adjacent splits make clear that this example does not by
itself imply that the learned boundaries coincide with linguistic units.

\section{Existing assets, licenses, and terms}
\label{app:assets}
Tables~\ref{tab:data-licenses} and~\ref{tab:model-licenses} report the repository identifiers
and license information verified from the official cards and repositories in August 2026.
The experiment records pin the cached revisions used. Assets with non-commercial or
unclear terms were used only for academic research and evaluation and are not redistributed.

\begin{table*}[ht]
  \caption{Datasets used in the reported experiments. ``Not specified'' means that the
    official repository does not publish a named license; it is not an inferred license.}
  \label{tab:data-licenses}
  \centering
  \scriptsize
  \setlength{\tabcolsep}{3pt}
  \begin{tabular}{p{0.22\linewidth}p{0.23\linewidth}p{0.47\linewidth}}
    \toprule
    Asset / configuration                  & Repository                            & Published license or terms                                                                                                 \\
    \midrule
    FineWeb \texttt{sample-10BT}           & \texttt{HuggingFaceFW/fineweb}        & ODC-By 1.0; use is also subject to Common Crawl's Terms of Use.                                                            \\
    English Wikipedia \texttt{20231101.en} & \texttt{wikimedia/wikipedia}          & CC BY-SA 3.0 and GFDL.                                                                                                     \\
    WikiText-103 raw v1                    & \texttt{Salesforce/wikitext}          & Card metadata lists CC BY-SA 3.0 and GFDL, while its license prose says CC BY-SA 4.0; both notices are retained.           \\
    LAMBADA OpenAI, English                & \nolinkurl{EleutherAI/lambada_openai} & Modified MIT license.                                                                                                      \\
    AG News                                & \texttt{fancyzhx/ag\_news}            & No named license; the card permits research and other non-commercial activity.                                             \\
    HotpotQA \texttt{distractor}           & \texttt{hotpotqa/hotpot\_qa}          & CC BY-SA 4.0.                                                                                                              \\
    MTEB evaluation package                & \nolinkurl{embeddings-benchmark/mteb} & Apache 2.0 code. Its STSBenchmark, STS17, and NFCorpus mirrors do not specify dataset licenses; SICK-R is CC BY-NC-SA 3.0. \\
    \bottomrule
  \end{tabular}
\end{table*}

\begin{table*}[ht]
  \caption{External model and baseline assets. License restrictions apply to the original
    assets; no third-party weights are redistributed with this preprint.}
  \label{tab:model-licenses}
  \centering
  \scriptsize
  \setlength{\tabcolsep}{3pt}
  \begin{tabular}{p{0.25\linewidth}p{0.29\linewidth}p{0.38\linewidth}}
    \toprule
    Asset                       & Repository                            & Published license or terms                                       \\
    \midrule
    Pythia-410M                 & \texttt{EleutherAI/pythia-410m}       & Apache 2.0.                                                      \\
    GPT-NeoX tokenizer          & \texttt{EleutherAI/gpt-neox-20b}      & Apache 2.0.                                                      \\
    Mamba2-130M and Mamba2-370M & \texttt{state-spaces/mamba2-*}        & Apache 2.0.                                                      \\
    Llama-3-8B                  & \texttt{NousResearch/Meta-Llama-3-8B} & Meta Llama 3 Community License and Acceptable Use Policy.        \\
    Qwen2.5-1.5B scorer         & \texttt{Qwen/Qwen2.5-1.5B}            & Apache 2.0.                                                      \\
    SONAR                       & \texttt{facebook/SONAR}               & MIT code; the text encoder and decoder weights are CC BY-NC 4.0. \\
    \bottomrule
  \end{tabular}
\end{table*}

\paragraph{Language-model use.} Language models are core experimental components: Pythia
and Mamba form the autoencoder, Pythia and Llama are readers, and Qwen is an external
perplexity scorer. Their roles and training are described in Sections~\ref{sec:method} and
\ref{sec:readout}. An LLM was also used for writing, editing, and formatting assistance;
it did not generate measurements or determine the experimental conclusions.

\end{document}